\documentclass[journal]{vgtc}                
\ifpdf
  \pdfoutput=1\relax                   
  \usepackage{graphicx}                
  \DeclareGraphicsExtensions{.pdf,.png,.jpg,.jpeg} 
\else
  \usepackage{graphicx}                
  \DeclareGraphicsExtensions{.eps}     
\fi%

\graphicspath{{figures/}{pictures/}{images/}{./}} 

\usepackage{microtype}                 
\PassOptionsToPackage{warn}{textcomp}  
\usepackage{textcomp}                  
\usepackage{mathptmx}                  
\usepackage{times}                     
\usepackage{cite}                      
\usepackage{tabu}                      
\usepackage{booktabs}                  
\usepackage{bm}
\usepackage{amsmath}

\onlineid{1294}

\vgtccategory{Research}
\vgtcpapertype{application/design study}

\title{Visual Neural Decomposition to Explain Multivariate Data Sets}

\author{Johannes Knittel, Andres Lalama, Steffen Koch, and Thomas Ertl}
\authorfooter{
\item
The authors are with the University of Stuttgart. \\
E-Mail: firstname.lastname@vis.uni-stuttgart.de
}

\shortauthortitle{A \MakeLowercase{\textit{et al.}}: Visual Neural Decomposition To Explain Multivariate Data Sets}

\abstract{
Investigating relationships between variables in multi-dimensional data sets is a common task for data analysts and engineers.
More specifically, it is often valuable to understand which ranges of which input variables lead to particular values of a given target variable.
Unfortunately, with an increasing number of independent variables, this process may become cumbersome and time-consuming due to the many possible combinations that have to be explored.
In this paper, we propose a novel approach to visualize correlations between input variables and a target output variable that scales to hundreds of variables.
We developed a visual model based on neural networks that can be explored in a guided way to help analysts find and understand such correlations.
First, we train a neural network to predict the target from the input variables.
Then, we visualize the inner workings of the resulting model to help understand relations within the data set.
We further introduce a new regularization term for the backpropagation algorithm that encourages the neural network to learn representations that are easier to interpret visually.
We apply our method to artificial and real-world data sets to show its utility.
} 


\CCScatlist{ 
 \CCScat{K.6.1}{Management of Computing and Information Systems}%
{Project and People Management}{Life Cycle};
 \CCScat{K.7.m}{The Computing Profession}{Miscellaneous}{Ethics}
}

\teaser{
  \centering
  \includegraphics[width=\linewidth]{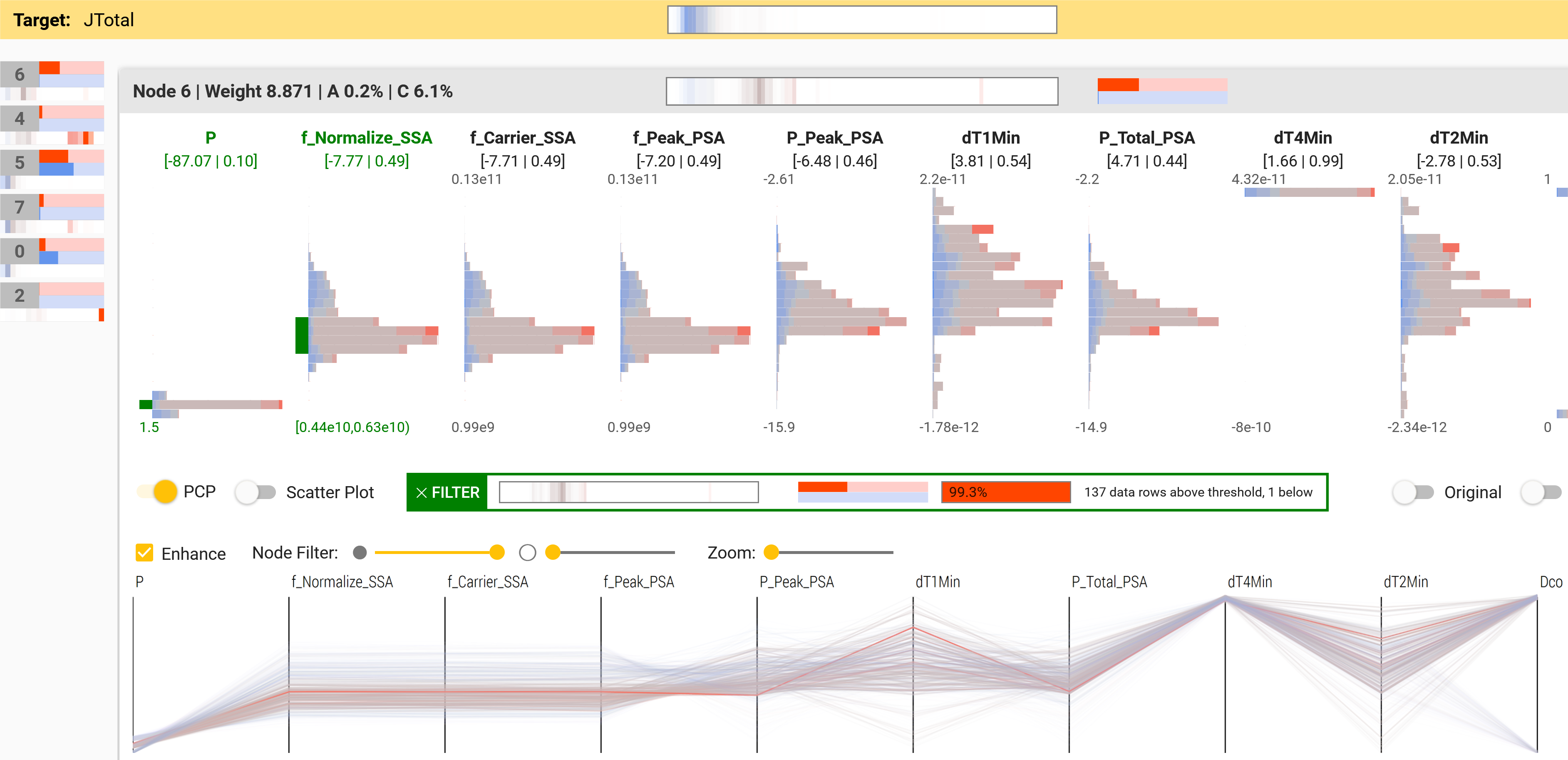}
  \caption{Visual Neural Decomposition of a chip testing measurement data set with the goal to identify cases in which the target variable (here: jitter) exhibits high values. Each node visualizes parts of the data set depending on its activation.}
  \label{fig:teaser}
}

\vgtcinsertpkg

\begin{document}


\firstsection{Motivation}

\maketitle

Many interesting real-world data sets have more than three dimensions, e.g., socio-demographic data to analyze consumer or voting behavior, meteorological measurements to predict the weather, or Integrated Circuit (IC) testing measurements to analyze the performance of chips.
Unfortunately, analyzing multivariate data sets is a challenging task, especially if the number of variables grows well above three.
One common goal analysts often face is to explore whether and how exactly the different variables influence a specific variable of the data set, the so-called \emph{target variable}.
For example, given a data set with background information on voters, can we detect patterns of groups that voted for a specific candidate?
To answer this question, we can use parallel coordinate plots (PCP)~\cite{Inselberg1985} or pick one or two variables at a time and visualize their relationship with the target variable individually, e.g., with a scatter plot matrix~\cite{Carr1987}.
However, such techniques do not scale well to many variables~\cite{Barlowe2008}, and analysts may fail to detect more complex relationships involving three or more variables, particularly if there are multiple groups of conditions that lead to high target values.
In case of the voting behavior data set example, a combination of a certain age range, level of education \emph{and} personal hobbies could explain \emph{one} cluster of specific votes, for instance.

Experts from a wide range of domains have to deal with such challenges, including manufacturing specialists like chip developers.
In post-silicon validation, produced IC chips are tested under varying conditions to detect design bugs and to reveal sensitivities of a target variable with respect to other variables as hints for design improvements~\cite{Mitra2010}.
We conducted a case study with a domain expert and research fellow of one of the world's largest provider of automatic test equipment to collect requirements for our approach.
The automated testing of the chips often produces large high-dimensional data sets with up to hundreds of variables that capture parameters, environmental factors, and resulting error measurements.
Engineers analyze the data sets to find out which parameter range combinations may lead to undesired behaviors of the chip (out-of-spec behavior), and to determine parameters for optimal performance (tuning), for instance, to maximize battery life of the chip.
While the total number of variables is high, the actual number of relevant variables for a specific case is often less than half a dozen, according to our expert.
Furthermore, most of the problems at hand can be formulated such that the goal is to find conditions under which a certain variable (e.g., error rate) has high values.
However, there can be several distinct patterns involving different variables that result in high values, for instance, several bugs relating to different environmental factors.
Some of these patterns may only appear in a small fraction of the data set.
As of now, the analysts and engineers make heavy use of scatter plot matrices to investigate possible correlations between several subsets of variables and the selected target variable, which is time-consuming.

Given a multivariate data set, the goal of our approach is to extract and visualize cases that lead to high values of a specified target variable.
The output of our model guides analysts to thoroughly explore interesting combinations of variables that correlate with the target.
Crucially, we aim to visually explain the behavior of the target variable with the original features, the remaining variables of the data set, because they often have a semantic meaning for analysts~\cite{Sedlmair2014}.
Thus, we need a visually interpretable technique that can recognize non-linear correlations between input variables and a specified target variable of the data set.
To achieve this, we first train a neural network using a novel architecture to predict the target variable based on the input of the other variables.
The neural network \emph{decomposes} the approximated function of the target variable into components, the \emph{neural} nodes in the hidden layer.
Then, we \emph{visualize} the behavior of these individual neurons to explain separately for each case in which conditions the target variable adopts high values.
We provide interactions to let analysts explore the found relationships in deep.
In other words, we do not apply visualization to explain an AI-approach (XAI~\cite{bertini_et_al:DR:2020:11982,Choo2018}), instead, we propose to use AI for visually explaining interesting relationships in data (AIX). 


We make the following contributions:
\begin{itemize}

\item We propose a novel model based on a neural network architecture to extract and visualize conditions in which a chosen target variable adopts high values.
It supports multi-dimensional and non-linear relationships, does not require analysts to pick interesting variables first, and can also recognize small-sized effects.
\item We present interactive node-specific parallel coordinate plots that leverage the trained weights of the neural network's hidden nodes to show a subset of the data set with a specific axis order.
\item We offer an integrated approach to load data sets, train models, and interactively explore and validate the visualized findings.

\end{itemize}






\section{Related Work}

Visual Neural Decomposition (VND) explains the behavior of a dependent variable (the target variable) with the remaining independent variables of the set, assuming that these variables have a semantic meaning for analysts (e.g., age).
This is similar to \emph{input-output models} in the context of visual parameter space analysis~\cite{Sedlmair2014}, but we do not aim to gain a systematic understanding of the complete input space, and depending on the use case and data set this is not even possible, for instance, regarding survey responses.
Godfrey and Gashler~\cite{Godfrey2017} introduced \emph{Neural Decomposition} as a new neural network technique to decompose time-series data into sums of periodic and nonperiodic components.
Similarly, we decompose the target variable into sigmoidal components based on the input variables, but we use classical neural networks with a novel regularization technique.
Our main idea is to \emph{visualize} the learned decomposition to understand which conditions lead to high values of the target variable.


\textbf{Multivariate data visualizations} such as scatter plot matrices~\cite{Carr1987} and parallel coordinates~\cite{Inselberg1985} are widely used to visualize multivariate data sets, but they do not scale well to many variables~\cite{Barlowe2008}.
Parallel coordinates have the advantage over scatter plot matrices that the required screen space only grows linearly with the number of variables, but the order of the axes, overplotting, and line-tracing are issues that have to be dealt with~\cite{Heinrich2013}.
There is ongoing research on how to improve parallel coordinates for large data sets, including drawing optimizations to reveal structures and avoid overplotting~\cite{Artero2004,Johansson2005,Novotny2006}, curve bundling to better reveal clusters~\cite{Luo2008}, and tuple-based statistics~\cite{Janetzko2016} as well as hierarchical aggregation~\cite{Richer2018} to make the visual analysis scalable.
Heinrich et al.~\cite{Heinrich2013} provide a more thorough overview of proposed PCP techniques and open challenges.

\textbf{Dimensionality reduction (DR)} methods, including MDS~\cite{Cox2000}, PCA~\cite{Abdi2010,Jolliffe2002}, t-SNE~\cite{VanDerMaaten2008}, and UMAP~\cite{McInnes2018} help to visualize multidimensional data items while still maintaining certain similarity measures between items, but the crucial question remains how analysts can relate findings to the original space.
To tackle this, Fujiwara et al. proposed ccPCA~\cite{Fujiwara2020} that clusters the result of the DR method and then visualizes for each cluster which features and feature ranges distinguish the respective cluster from the others the most.
Similarly to our method, the cluster-specific ranking of the features and corresponding histograms support the analysis of data sets with many variables.
However, the goal of VND is slightly different, in that we want to visualize which features influence a specific target variable the most.
In addition, VND calculates the (soft) clustering in the original high-dimensional space. 
Gleicher's approach~\cite{Gleicher2013} projects items onto user-defined, interpretable dimensions, e.g., a linear combination of certain variables.
Embeddings based on generalized barycentric coordinates such as RadViz Deluxe~\cite{Cheng2017} preserve the relation to the original space to some extent, but they work best if the data items have a dominant dimension.


\textbf{Explorative and iterative approaches} help to make the analysis of large multivariate data sets scalable.
Voyager \cite{Wongsuphasawat2016,Wongsuphasawat2017} supports open-ended and focused exploration of multivariate data sets with automated recommendations and interactive chart specifications.
DICON~\cite{Cao2011} visualizes multidimensional clustering results for cluster interpretation and comparison.
Scherer et al.~\cite{Scherer2011} introduced regressional feature vectors to enable visual sketch-based queries and to explore interesting scatter plots.
Behrisch et al.~\cite{Behrisch2015} propose a feedback-driven approach that iteratively learns the preferences of the user to make valuable suggestions for further explorations.
Turkay et al.~\cite{Turkay2012} present a method to interactively generate representative factors that combine several data points across dimensions in order to reduce the number of dimensions.
For geospatial data sets, specific methods were developed that facilitate maps~\cite{Goodwin2016,Malik2012}.
Several approaches rank variables and pairs of variables according to statistical correlation factors~\cite{Eichner2019,Malik2012,Piringer2008} or classification metrics such as separability~\cite{Tatu2009}.
Barlowe et al.~\cite{Barlowe2008} visualize partial derivatives of the dependent variable with regard to the independent variables and analysts can iteratively explore multi-correlations.
SmartStripes~\cite{May2011} ranks and visualizes correlations based on \emph{partitions} of the input features.
Analysts can iteratively select multiple partitions to explore more complex relationships.
Zhang et al.~\cite{Zhang2015} developed the correlation map to visualize pairwise correlations between variables in a graph.
The work of Klemm et al.\cite{Klemm2016} visualizes correlations of up to three variables with regard to a target variable in a 3D cube.
Our proposed system can recognize and visualize non-linear correlations as well as correlations involving more than three variables.

\textbf{Model building and partitioning} play an important role in several approaches.
INFUSE~\cite{Krause2014} supports interactive model building by visualizing the predictiveness of features according to several feature selection algorithms.
An integrated visual analytics approach to build logistic regression models was introduced by Zhang et al.~\cite{Zhang2016}.
Dingen et al.~\cite{Dingen2019} argue that building such models should be an iterative process selecting variables one at a time, because clinicians, for instance, need sparse and justified models. They developed RegressionExplorer to let analysts build and compare different regression models.
The approach of Muehlbacher and Piringer~\cite{Muhlbacher2013} allows to iteratively define and validate regression models.
They partition the input features into bins and visualize the target variable over binned features and pairs of features in a heatmap, ranked by their usefulness in predicting the target variable.
Bernard et al.~\cite{Bernard2014} developed a system that operates on partitions of the input space as well, but with the goal to find multivariate relations between specific bins across attributes.
They focus on detecting conditions that are statistically significant.
Analysts can select one or several bins to reveal associated clusters with related bins based on pairwise mutual information.
Such binning (or partitioning) strategies make the visual analysis of large data sets scalable through aggregation.
Another advantage is that even linear models can express certain non-linear relationships if the atomic unit is a bin (i.e., value range) and not a global variable anymore.
While uni- or bivariate metrics (e.g., Pearson correlation or mutual information) to rank features or cluster data items can offer statistically sound results, they bear the risk that analysts miss more complex multi-correlations.
For instance, there are cases in which only a combination of multiple variables may explain a specific behavior of the target variable, but looking at the individual variables or pairs of variables separately would not reveal a pattern.

\textbf{Subspace extraction} is a promising strategy to reduce the number of dimensions that have to be processed (e.g., for subsequent clustering methods) and visualized~\cite{Assent2007,eurova.20171118,Gunnemann2010,Hund2016,Jackle2018,Krause2017,Muller2008,Tatu2012,Tatu2012a,Wang2018}.
Tatu et al.~\cite{Tatu2012} employ an algorithm that detects interesting subspaces which are then grouped by similarity.
Parallel coordinate and scatter plots display the data set in each subspace.
SubVis~\cite{Hund2016} uses the OpenSubspace framework~\cite{Muller2011} to first extract subspaces and then find clusters within these subspaces.
They apply Multidimensional Scaling to provide a visual overview of all clusters, from which analysts can select similar subspace clusters.
The aggregation table visualizes the distributions of all related dimensions of a particular cluster.
SeekAView~\cite{Krause2017} supports building and refining subspaces interactively, including suggestions for interesting dimensions, e.g., based on how useful the respective variable is for predicting a target variable.
Our method applies a global approach for finding interesting patterns and we do not extract fixed subspaces, but analysts can use the range filter (Section~\ref{sec:rangeFilter}) to define a subspace (and a cluster within that subspace) based on the most important variables of a node.

\textbf{Decision trees} are commonly used to visualize different outcomes of the target variable depending on different splits of the input variables.
BaobabView~\cite{VanDenElzen2011} allows analysts to interactively construct and refine such decision trees.
Muehlbacher et al.~\cite{Muhlbacher2018} focus on building pareto-optimal decision trees as a trade-off between accuracy, complexity, and interpretability.
Decision trees are a popular choice because they are usually easy to comprehend and interpret.
However, continuous relations are difficult to model (e.g., relationship between horsepower and acceleration time) and slightly more complex patterns and 'splits' (e.g., two distinct ranges of a variable) can lead to overly complex trees, which makes it difficult for analysts to trace single paths.

\textbf{Visualizing (deep) neural networks} has become a popular research focus in recent years.
However, in most cases the approaches focus on explaining and debugging the \emph{models}, for instance, visualizing which pixel regions are most supportive for the prediction~\cite{Zintgraf2019}, explaining predictions of convolutional neural networks with surrogate decision trees~\cite{Jia2020}, or visualizing activation patterns to understand deep learning models~\cite{Kahng2018}.
Liu et al.~\cite{Liu2017a} provide an overview of visual analytic approaches for understanding, debugging and refining machine learning models, Choo et al.~\cite{Choo2018} for explainable deep learning, and recently Sacha et al.~\cite{Sacha2019} for assisting machine learning.
Conversely, we use neural networks to understand the relationships of the \emph{underlying data}.
Nevertheless, some approaches indirectly also reveal structures of the data.
CNNVis~\cite{Liu2017a} visualizes the learned features of neurons and their interactions to analyze image-based CNNs.
While the aim is to refine and debug such networks, the resulting visualizations also offer a glimpse into the structure of the training data set, including prevalent features of the images and different clusters (e.g., cats and dogs).
Likewise, LSTMVis~\cite{Strobelt2018} allows analysts to retrieve similar sentences and paragraphs in the corpus, even though the approach is about visualizing hidden states of recurrent neural networks.

\section{Method}

Given a multivariate data set, we want to investigate in which cases a specified target variable $y$ adopts values in a specified range.
The goal of our method is to visualize such cases separately to provide digestible visual explanations, while still supporting more complex relationships between $y$ and the remaining variables of the data set.

Multivariable linear regression is often used to infer which independent variables $x_i$ seem to have the most influence on the dependent variable $y$ by solving $y=\sum_i{a_i x_i} + b$ for the coefficients $a_i$~\cite{Schneider2010}.
While the resulting coefficients are easy to interpret, linear regression cannot capture more complex correlations involving non-linear relationships or combinations of variables (e.g., $y$ is only high if two input variables are both high).
Neural networks, on the other hand, can approximate highly complex functions. 
In simplified terms, a single-output neural network essentially combines several \emph{localized} quasi-regressions as represented by the hidden nodes.
We assume that some of these nodes capture specific cases that lead to values of $y$ in the desired range.
Hence, the main idea of our approach is to first train a neural network predicting the target based on the other variables in the data set (\emph{input variables}).
Afterward, we use the hidden nodes' activations to visually explain how certain parts and variables of the data set correlate with the target.
We introduce a novel regularization technique (Section~\ref{sec:regularization}) that encourages the neural network to model the relationships in a way that is easier to interpret.
In other words, we use the training process to perform multiple non-linear regressions, and analysts can then further explore these in our interactive visual interface.

We are mainly interested in how our model captures the underlying dynamics, thus, the resulting accuracy of our network is less relevant.
However, a consistently low accuracy indicates that there is either a very subtle or no relationship between inputs and output, because neural networks can approximate complex, non-linear functions (a particular training run may not converge nevertheless).
We do not aim at debugging machine learning models to improve the model or find weak spots, we rather analyze them visually to understand the underlying data sets they were trained on.
In this paper, we focus on explaining high target values for simplicity, but this does not limit the range of our approach.
If analysts are interested in low- or mid-value cases, $y$ could be transformed accordingly to fit our definition (e.g., $\hat{y} = -y$ for low-value cases).

\subsection{Basic Architecture}
\label{sec:architecture}

\begin{figure}
  \centering
  \includegraphics[width=0.7\linewidth]{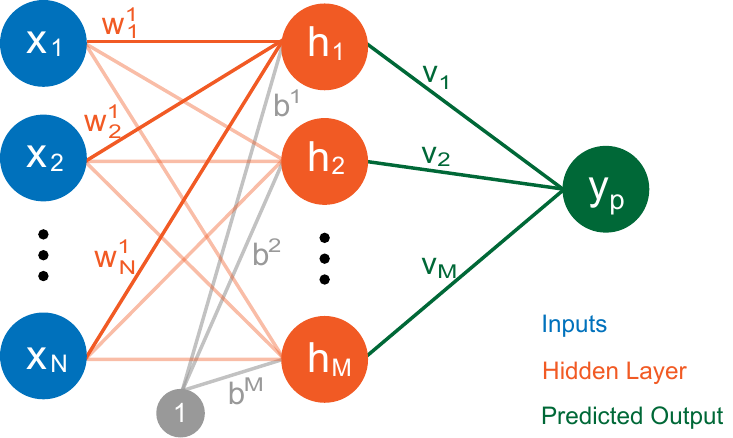}
  \caption{Architecture with inputs $\vec{\bm{x}}$, input weights $\vec{\bm{w}^i}$, biases $b^i$, hidden node outputs $h_i$, hidden node weights $v_i$, and prediction $y_p$ of the target.}
  \label{fig:nnModel}
\end{figure}

Figure~\ref{fig:nnModel} depicts the architecture of our model.
We use a fully-connected, feed-forward neural network with one hidden layer and one output node.
To compute the (scalar) output $h_i$ of a hidden node $i$ in an artificial neural network, the dot product of the inputs $\vec{\bm{x}}$ with the trained weights of the node $\vec{\bm{w^i}}$ plus a constant offset $b^i$ (the bias) is fed into a non-linear, monotonically increasing activation function $\sigma (z)$:

\begin{equation}
h_i(\vec{\bm{x}}) = \sigma( \vec{\bm{w^i}} \cdot \vec{\bm{x}} + b^i )
\end{equation}

We use the sigmoid function $\sigma(z)=\frac{1}{1+e^{-z}}$ as activation function which returns values between zero and one.
If the function output is close to zero, we say that the respective hidden node is \textit{inactive}, otherwise, it is \textit{active} (to a certain degree).
Similar to linear regressions, inputs with positive associated weights (or coefficients) have a positive monotonic relationship with the output of the hidden node (i.e., higher values lead to equal or higher outputs), inputs with negative weights have a negative relationship, and inputs with weights close to zero do not significantly impact the output.
However, artificial neural networks are more powerful than linear regressions, because they use non-linear activation functions.
In simple terms, the activation function enables thresholds, i.e., the node's output may only start to grow significantly above zero if the weighted sum reaches a certain value, and it nearly stops to grow once it is close to one.
Apart from this non-linearity, the relation of the hidden node output to its inputs as defined by the weights is still monotonic and smooth, which we exploit in our approach.
Understanding the behavior of individual hidden nodes is therefore easier than understanding the model as a whole.
The final output $y_p(\vec{\bm{x}})$ is assembled from the outputs of the hidden nodes as weighted sum:

\begin{equation}
y_p(\vec{\bm{x}}) = \sum_i{h_i(\vec{\bm{x}}) v_i}
\end{equation}

In the following, we call hidden nodes with a positive associated weight $v$ in the final layer \emph{positive nodes} (they drive the final prediction up), and nodes with a negative $v$ in the final layer \emph{negative nodes}.
In contrast to a typical feed-forward architecture, we do not use the bias in the final layer to make sure that at least one positive node has to be active for high target values.
We scale every variable (inputs and output) so that they are in the range $[0,1]$.
The lack of bias and the normalization of the output ensures that the resulting prediction of the neural network is (only) high if at least one positive node is sufficiently active and negative nodes are mainly inactive.
Hence, positive nodes may offer hints which variables and conditions correlate with high target values.

\subsection{Hidden Node Filtering}
\label{sec:nodeFiltering}


In our approach, each positive node acts as a filter.
We assume that during training, some hidden nodes capture specific cases that lead to a high target value (\textit{high-target cases}), so filtering the data according to which hidden node is active can offer insights into the cases we are interested in.
This decomposition enables analysts to study the relationships between input variables and target separately for distinct conditions.
In particular, the input weights of a node indicate how important each input is for the computation of this node, and analysts can focus on the most promising variables even if the total number of inputs is rather high.

However, for our analysis goal, it is often not enough to filter the data items solely based on whether a particular hidden node is active (i.e., its output $h_i(\vec{\bm{x}})$ is high).
Say, for instance, our target variable peaks if the (single) input is in the middle at $0.5$, but it is low if the input is near zero or one.
One hidden node alone cannot model this case, because due to the monotonic nature of the activation function the output of the node cannot go down again after the input has reached $0.5$.
The neural network could use a second \emph{negative} node to approximate this function, i.e., the positive node models the ramp-up to $0.5$ and the negative node the ramp-down to zero again.
We can say that the second, negative node \textit{inhibits} the output of the positive node if the input is above $0.5$.
Therefore, we have to take into account the effects of such \emph{inhibitors} while a particular hidden node is active.
To achieve this, we visualize the data distributions of the inputs and the target output for such data items where the respective hidden node of interest is \emph{contributing} and not just active, which is a smaller subset.

Regarding a data item, we say a node is \emph{contributing} if this node is active \emph{and} the weighted output of the node is high compared to all other weighted outputs of positive nodes \emph{and} the prediction is high.
In other words, we want to detect \emph{salient} stimulations of hidden nodes by the input data that travel through to the final output value.
More formally, let $V$ be the set of hidden nodes, then the contribution $c_i(\vec{\bm{x}})$ of hidden node $i$ regarding data item $\vec{\bm{x}}$ is:

\begin{equation}
c^i(\vec{\bm{x}}) = \frac{1}{Z} h_i(\vec{\bm{x}}) \frac{\min (y_p(\vec{\bm{x}}), h_i(\vec{\bm{x}}) v_i )}{\sum_j{h_j(\vec{\bm{x}}) v_j} }, \quad \forall j \in V: v_j > 0
\end{equation}

We take the final prediction if it is lower than the weighted output of the node to ensure that the node is not suppressed by other (negative) nodes.
$Z$ is a scaling factor to retrieve a contribution of $1$ for data items that stimulate the particular node the most.
The advantage of filtering by contributions instead of just activations is that the resulting 'clusters' can model highly non-linear relationships.
For instance, one cluster could model the case that our target variable correlates linearly with $a$ \emph{if} variables $b$ and $c$ are both within a specific range.

\subsection{Ranking of Variables}
\label{sec:ranking}

The filtering (or clustering) helps analysts to focus on specific parts of the data set, but without a suitable ranking of the variables, it may still be tedious to gain insights into such clusters, particularly if the total number of variables is high.
We want to rank the variables according to their impact on the contribution of a particular hidden node.
As explained in Section~\ref{sec:architecture}, the output of a hidden node is monotonically related to the weighted sum of all inputs.
On the one hand, this means that the magnitude of the weight is an indicator of the impact of the corresponding variable.
On the other hand, we also have to take the distribution of the variable into account.
A high positive weight is less important if the average value of the variable is close to zero, for instance.

For an input $k$ of node $i$ with a positive correlation (i.e., $w_k^i > 0$), we calculate the average value whenever the node $i$ is contributing and multiply it with the weight $w_k^i$.
This determines which variable on average has a high share of the weighted sum while the node is contributing, and is therefore a driving force of high activations.
For an input $l$ of node $i$ with a negative correlation (i.e., $w_l^i < 0$), we determine the average value whenever the node $i$ is \emph{not} contributing and multiply it with the absolute value of the weight $|w_l^i|$.
The resulting value tells us which variable has the highest impact on inhibiting the output whenever the node is not contributing, and is therefore enabling high activations if the values are low.
More formally, let $X$ be the input rows of the data set. We define the rank $r_k^i$ of the variable $k$ regarding hidden node $i$ as follows:

\begin{equation}
\begin{split}
r_k^i =  \frac{|w_k^i|}{S} \sum_{\vec{\bm{x}} \in X}{ x_k \, \hat{c}_k^i(\vec{\bm{x}}) }, \quad S = \sum_{\vec{\bm{x}} \in X}{\hat{c}_k^i(\vec{\bm{x}})} \\
\hat{c}_k^i(\vec{\bm{x}})= 
\begin{cases}
    c^i(\vec{\bm{x}}), & \text{if } \ w_k^i > 0  \\
    1- c^i(\vec{\bm{x}}), & \text{otherwise}
\end{cases}
\end{split}
\end{equation}

\subsection{Homogeneous Regularization}
\label{sec:regularization}

We want to leverage machine learning to decompose the structure of the target variable into separate components depending on the input variables to explain in which cases the target variable is high.
Sometimes, the resulting nodes still combine several cases that we would have liked to be represented separately, so that analysts can easily interpret the relations visually.
We therefore introduce a regularization technique that encourages positive hidden nodes to be mainly active in high-value predictions, and negative nodes (inhibitors) to be mainly active in low-value predictions, i.e., we want a more \emph{homogeneous} behavior of individual nodes with respect to their activation pattern.
We say a hidden node $i$ is \emph{positive}/\emph{negative} if the associated weight $v_i$ to calculate the final prediction is positive/negative.

Let $\tau$ be the threshold that defines the border between low- and high-value cases of our target variable.
For each data item $\vec{\bm{x}}$ and hidden node $i$ in the batch, we define:

\begin{equation}
    a_i(\vec{\bm{x}}, v_i)= 
\begin{cases}
    1, & \text{if } \; y(\vec{\bm{x}}) \geq \tau \land v_i < 0 \;\; \lor \;\; y(\vec{\bm{x}}) < \tau \land v_i > 0  \\
    0, & \text{otherwise}
\end{cases}
\end{equation}

In other words, $a_i(\vec{\bm{x}}, v_i)$ is $1$ if the current node $i$ is a positive node and the current target value is below the threshold, or it is $1$ if the current node is an inhibitor and the current target value is above the threshold.
Let $V$ be the set of hidden nodes.
Our loss function $L$ for a single data item $(\vec{\bm{x}}, y)$ with the weights $W$ and biases $B$ is then defined as

\begin{equation}
\begin{split}
L(\vec{\bm{x}}, y, W, B) = & \frac{1}{2} {\left(  \left(\sum_{i \in V}{  h_i(\vec{\bm{x}}) v_i} \right) - y(\vec{\bm{x}}) \right)}^2 + \frac{1}{2} \beta \sum_{i \in V}{ a_i(\vec{\bm{x}}, v_i) h_i(\vec{\bm{x}})^2}
\end{split}
\end{equation}

The first part of the loss function is just the typical squared loss to let the network learn accurate predictions.
The second part is our homogeneous regularization term that penalizes high node activations of positive nodes for data items that ultimately result in low final outputs and high node activations of negative nodes for data items that result in high final outputs.
The hyper-parameter $\beta$ defines the strength of our regularization.

In general, the goal of any regularization is to train a model that is better according to some criteria (e.g., generalizability to unseen data) by limiting the solution space.
In our case, we want to ensure that the neural network models the task in a `simpler' way such that the visualized behavior of individual nodes is easier to comprehend.

For illustrative purposes, let us assume we have two input variables \textit{A} and \textit{B}, and the target variable \textit{Y} is high if (and only if) \emph{one} of the two input variables is high and the other \emph{low}.
A neural network could model the target function as follows: one hidden node is active if the sum $A+B$ is sufficiently high, and the other hidden node with a negative weight is active if the sum $A+B$ is \emph{higher} than the maximum possible value of either input.
In this case, the first node captures \emph{both} high-value scenarios (A is high and B is low, as well as B is low and A is high).
However, the first node is \emph{also} active in a low-value case, namely if A and B are both high.
The final prediction is only low in this case because the second node with the negative weight inhibits the output of the first node if $A$ and $B$ are both high.
While such a network makes perfectly fine predictions, it is more difficult to interpret visually, because one node captures both high-value scenarios.
The proposed regularization, though, avoids this constellation.
The network is encouraged to model the relationship in a different way with a greater specialization of each node.
One hidden node would \emph{only} be active if A is high and B low, and the other if B is high and A low, because this leads to a lower loss of our regularization term (while still predicting equally well).
The visual representation of the resulting model can then easily explain the two conditions separately.

We performed a hyper-parameter analysis on a synthetic data set with different values for $\beta$ and several network sizes.
The results show that our homogeneous regularization ($\beta \geq 0.1$) helps to extract all high-values cases, particularly the more subtle ones.
A higher number of hidden nodes also increased the likelihood of detecting the patterns we were interested in.
However, the results also show that our approach is generally not very sensitive regarding the hyper-parameters.
Appendix A in the supplemental material contains detailed results of our analysis.

We train the network with mini-batch gradient descent and RMSprop to iteratively derive the weights and biases.
After training, we compute for all data items the respective contribution of each hidden node and rank the variables for each node.
We use the contributions of a particular node to filter the data items for the histogram and parallel coordinate plots that we describe in detail in Section~\ref{sec:design}.

\section{Application Design}
\label{sec:design}

Our approach extracts and visually explains multivariable correlations in a data set regarding a specified target variable.
After performing the neural decomposition as described in the previous section, we visualize the distributions of the input variables in relation to our target variable separately for each positive hidden node in our model, based on the calculated contributions (Section~\ref{sec:nodeFiltering}).
This helps analysts to understand which variables and ranges correlate with high target values, as each cluster may represent a different group of conditions.
With \textit{target variable} we always refer to the actual values in the data set, not the predictions of the model.
We only use the predictions to derive the clustering (Section~\ref{sec:nodeFiltering}).
We developed an integrated workflow that enables analysts to load data sets, configure variables, set parameters, train the model, visualize the results, and interact with the views, all within one application.
A typical workflow starts with selecting the target variable of the loaded data set.
The analyst can then train a neural network that tries to predict the target value based on the other variables.
After training, we visualize the resulting clustering with stacked histograms for each hidden node.
They offer a compact visual representation of the conditions that lead to high target values.
The histograms show the distribution of the variables for a subset of data items that stimulate the particular node.
Analysts can then explore and verify the found correlations in a targeted manner with interactive, node-specific parallel coordinate plots and by using simple range filters.

\subsection{Variables}

\begin{figure}
  \centering
  \includegraphics[width=\linewidth]{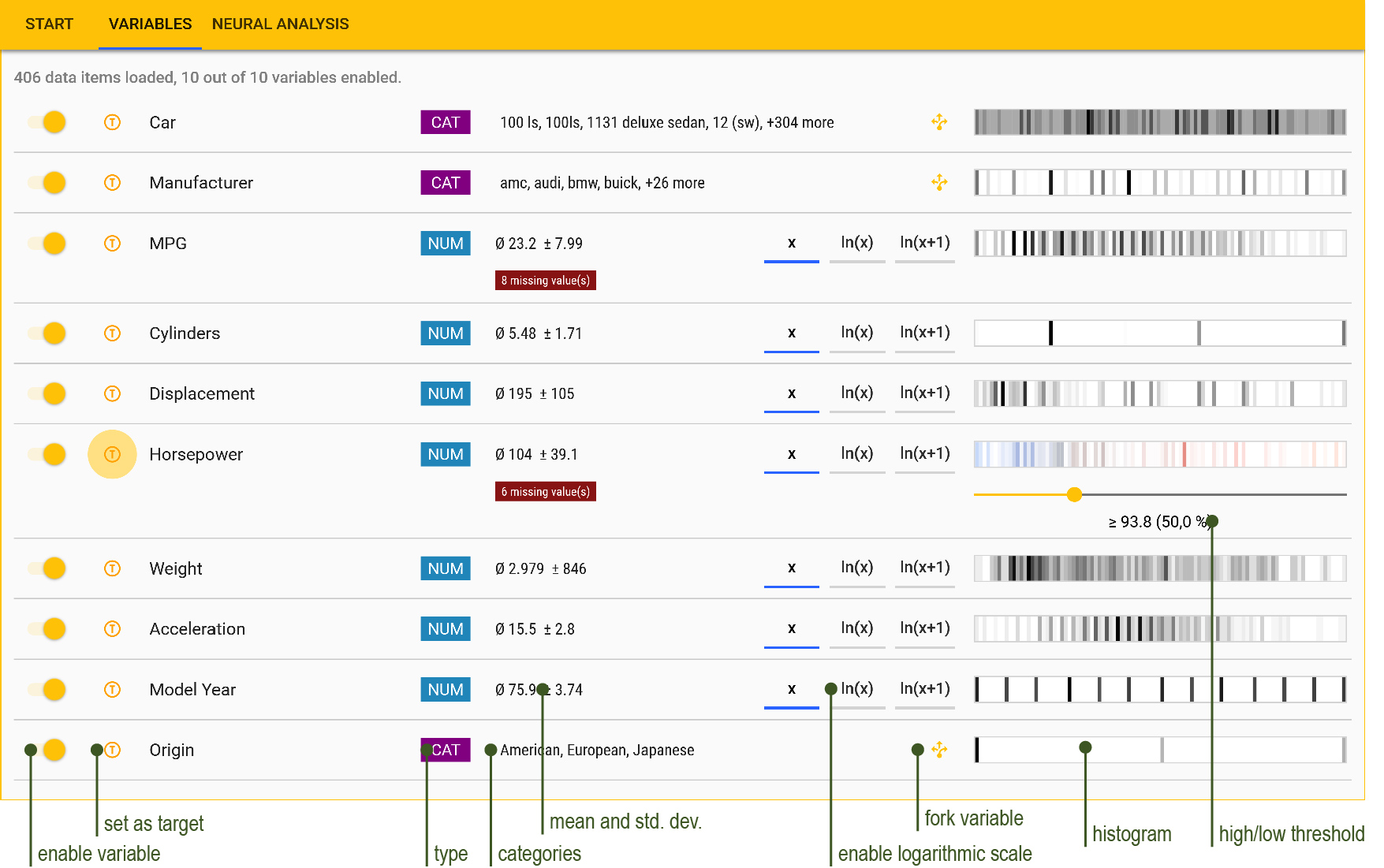}
  \caption{All available variables of the automobiles data set with previews of the data, scale, and histogram.}
  \label{fig:cars-variables}
\end{figure}

After loading the data set, the detected variables are listed in the respective tab as shown in Figure~\ref{fig:cars-variables}.
Here, the 1983 Data Exposition data set~\cite{Ramos1983} has been loaded that contains a list of (rather ancient) automobiles with several properties and technical specifications, e.g., the horsepower of the car and its driving range in miles per gallon.
Analysts can choose which variables should be part of the subsequent analysis process with the toggles on the very left of each row.
To select the target variable, they can click on the \textit{(T)} button next to the name of the respective variable.
In Figure~\ref{fig:cars-variables}, \textit{Horsepower} was selected as target variable, i.e., the analyst wants to explore how the other variables relate to the horsepower of the car.

Upon loading, our application automatically analyzes the given data set to determine the type of each variable (categorical or numerical) and whether a logarithmic scale should be applied.
We encode each category as a number between zero and one.
While this embedding allows to represent hundreds of categories with just one variable, in machine learning it is often beneficial to split categorical variables into distinct binary variables that represent whether the specific category is \emph{on}, i.e., whether it is the current value of the original variable.
This is also necessary if we want to designate a specific category as target.
Analysts can click on the yellow \textit{fork} button left to the histogram (in Figure~\ref{fig:cars-variables}) to generate said distinct variables for each category.
Numeric inputs are scaled to the range from zero to one, after having transformed them to the logarithm if needed.
Missing values are replaced with the respective regular minimum ($0$ after scaling) to avoid distortions.
The middle of each row displays the type of the variable alongside statistical properties such as the number of categories or mean and standard deviation.
Next to it, analysts can override the automatic decision whether a logarithmic scale should be applied or not.

To the right, a histogram provides insights into the distribution of the variable.
A darker shade indicates a higher number of values around that region.
For the selected target variable, a slider underneath the histogram appears which allows analysts to define a custom threshold.
The percentage in brackets indicates how many data rows fit this threshold and count as high-value cases.
The goal of our approach is to visually explain in which conditions the target variable is higher than said threshold.
The middle value between maximum and minimum of the respective variable is the default, but in Figure~\ref{fig:cars-variables}, the analyst has changed it to the median.

\subsection{Model Training}

\begin{figure}
  \centering
  \includegraphics[width=\linewidth]{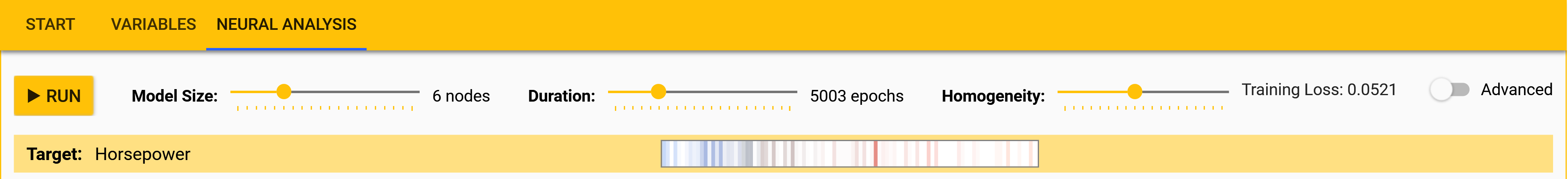}
  \caption{Meta-parameters of the neural network training, e.g., complexity of the model in terms of the number of hidden nodes, or duration in terms of the number of iterations.}
  \label{fig:cars-parameters}
\end{figure}

In the third tab (\textit{Neural Analysis}), analysts can start the training of the model.
Figure~\ref{fig:cars-parameters} shows how analysts can modify the model size, duration, and the strength of the regularization with sliders located at the top.
Bigger models may better capture small-sized effects in the data, but take longer to train and can lead to some redundant nodes showing mostly the same information.
The resulting error on the training set is depicted at the top-right corner after each run.
Underneath the sliders, we show again the histogram of the selected target variable, which is helpful for comparison with the subsequent node visualizations.

\subsection{Node Visualization}

\begin{figure}
  \centering
  \includegraphics[width=\linewidth]{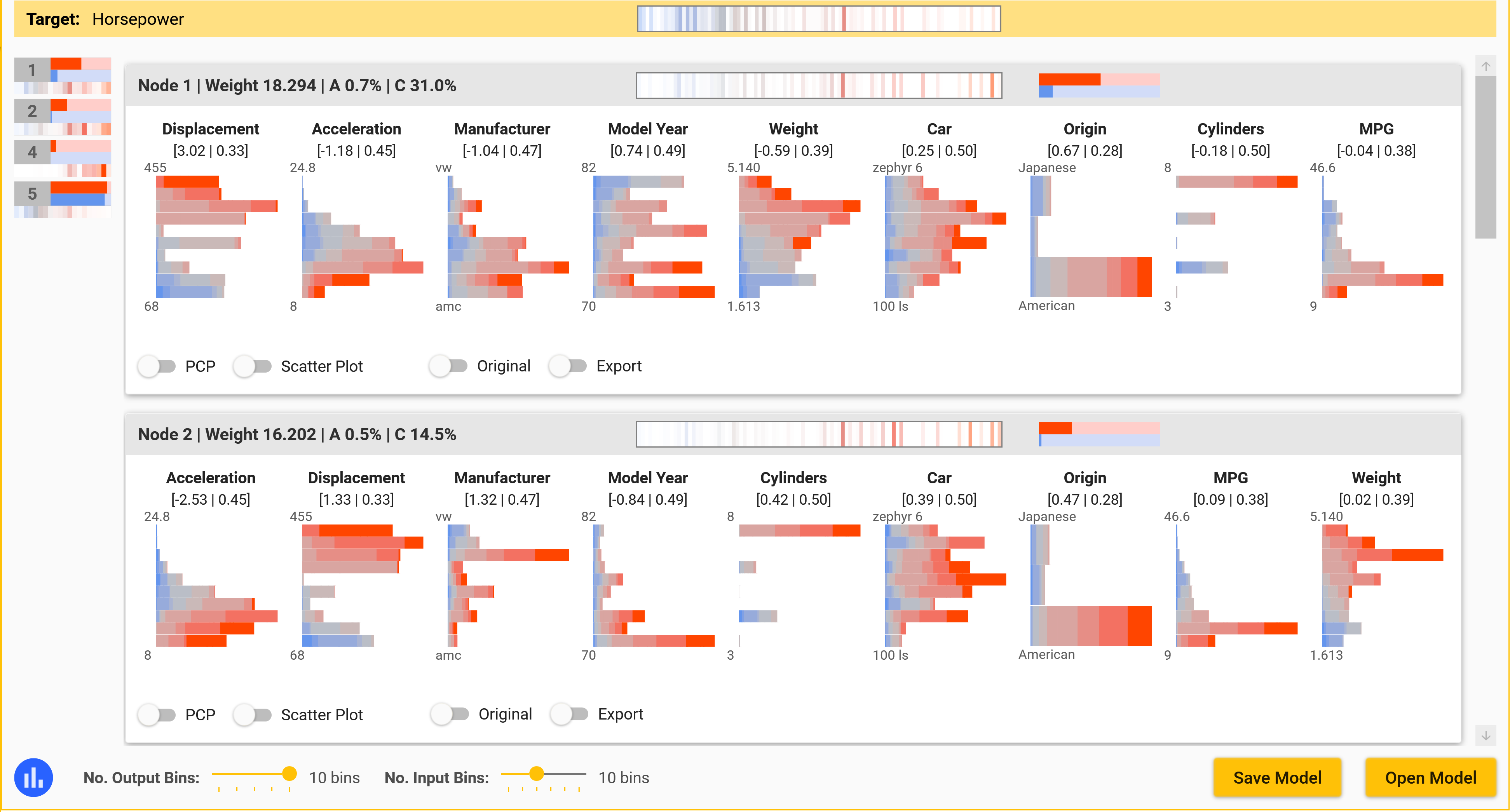}
  \caption{Visualization of the resulting model. Each card represents one hidden node and visualizes the data set filtered by the contribution of the  respective node. A compact overview of the nodes is shown on the left.}
  \label{fig:cars-nodes}
\end{figure}

\begin{figure}
  \centering
  \includegraphics[width=0.7\linewidth]{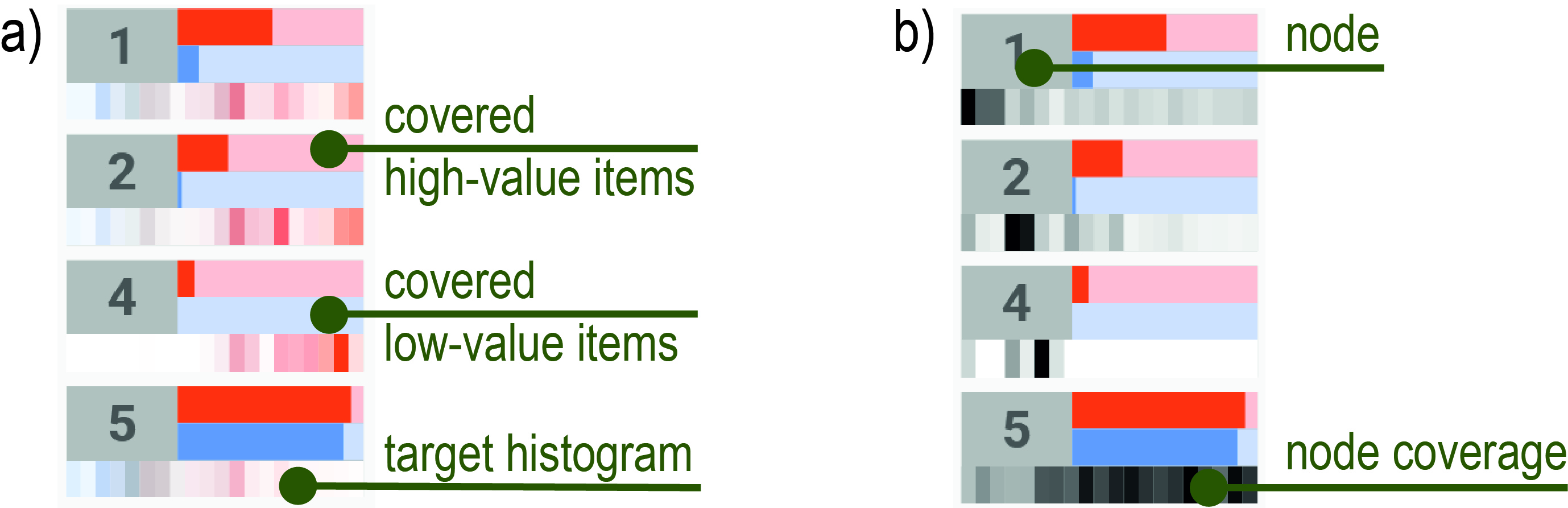}
  \caption{Compact overview of all positive nodes: a) node-specific histogram of target, b) node-specific coverage of high-value cases}
  \label{fig:compact-overview-explained}
\end{figure}


We want to investigate under which conditions our target variable adopts values above a certain threshold (\emph{high-value cases}).
We call data items with a target value above the threshold \emph{high-value items}.
Our approach assumes that some of the positive hidden nodes in our model specialize in particular high-value cases that can then be explained visually.
We distinguish between high- and low-value items and cases to improve the utility of our visualization, and it also influences the homogeneous regularization (Section~\ref{sec:regularization}).
For instance, we generally associate high-value items with red, and low-value items with blue colors.
However, the neural network learns to predict the actual target value irrespectively of the threshold.

Figure~\ref{fig:cars-nodes} shows the resulting visualizations of the positive nodes after training.
We introduce for each node-defined cluster a \textit{card} that shows the distribution of the data subset the respective node contributes to.
We chose the card-based layout because it maps well to our neural decomposition, in that each node represents one distinct cluster of high-value cases that analysts can explore separately, with aggregated statistics for each case.
At the same time, it also enables the comparison between nodes by presenting the cards in a list, with the possibility to rank nodes by their relevancy for the problem at hand.

Each card presents stacked histograms (Section~\ref{sec:histograms}) of the input variables, ordered by the importance of the respective variable for the cluster as described in Section~\ref{sec:ranking}.
They offer a visual summary for which input ranges the node is contributing in relation to the target values.
We hide variables with a very low importance score (5\% of the first, most important variable per default).
The actual coefficient (weight) and average activation of each input is displayed in squared brackets below the name of the variable.
The gray header displays the weight, average activation (A), and average contribution (C) of the node.
The red and blue bars to the right indicate how many of the high- and low-value items the node covers.
For instance, if the red bar is at 100\% then this means that the node is contributing to all data items with a target value above the threshold.
In the ideal case, the blue bars should be rather small, because if red and blue bars are both strong the respective cluster covers both low- and high-value cases.

In the middle of the gray header, a small histogram depicts the distribution of the target value whenever the node is contributing (target histogram). 
Users can switch to display the black \emph{node coverage} instead of the target histogram using the blue toggle button on the bottom left of the window (Figure~\ref{fig:cars-nodes}).
While the target histogram visualizes on which regions of the \emph{target variable} the node specializes, the node coverage  instead shows for which high-value \emph{data items} the node is contributing.
This helps to understand which nodes complement one another and which cover similar data items.
To compute the node coverage, we take each high-value item, sort them by which node contributes most on the respective item, assign each a running number, and then build a histogram of all the respective running numbers where the particular node is contributing, i.e., each data item is represented by a vertical line colored from white to black depending on the contribution.
On the left side of the window (Figure~\ref{fig:cars-nodes}), the node summary provides analysts an overview of all nodes with compact versions of the histograms and case coverages that are displayed in the respective headers. 
Depending on the setting, either the target histogram (Figure~\ref{fig:compact-overview-explained} a)) or node coverage (Figure~\ref{fig:compact-overview-explained} b)) is shown.

The total number of nodes can be high.
To help users focus on promising ones we show those nodes on top which explain high-value cases.
We realize this ranking by multiplying the number of affected high-value items of the node with the logarithm of the total number of low-value items divided by the number of affected low-value items.
Taking the logarithm of the fraction ensures that we boost nodes covering many high-value items, but \emph{relatively} few low-value cases.

The target histograms in Figure~\ref{fig:cars-nodes} of Node 1 and 2 are similar, so the nodes cover items with a similar distribution of the target variable.
The node coverage in Figure~\ref{fig:compact-overview-explained} b) reveals, however, that Node 2 contributes most to different data items than Node 1.
\textit{Displacement} has the most impact on Node 1 with a weight of $3.026$, followed by \textit{Acceleration}.
The signs of the weights indicate a positive correlation for the Weight (bigger cars in the data set seem to have more horsepower) and a negative correlation with Acceleration (faster acceleration times seem to coincide with more horsepower).
It should be noted that these node-specific correlations do not necessarily mean that there is an overall statistical correlation with the target variable.

\subsubsection{Stacked Histograms}
\label{sec:histograms}

\begin{figure}
  \centering
  \includegraphics[width=\linewidth]{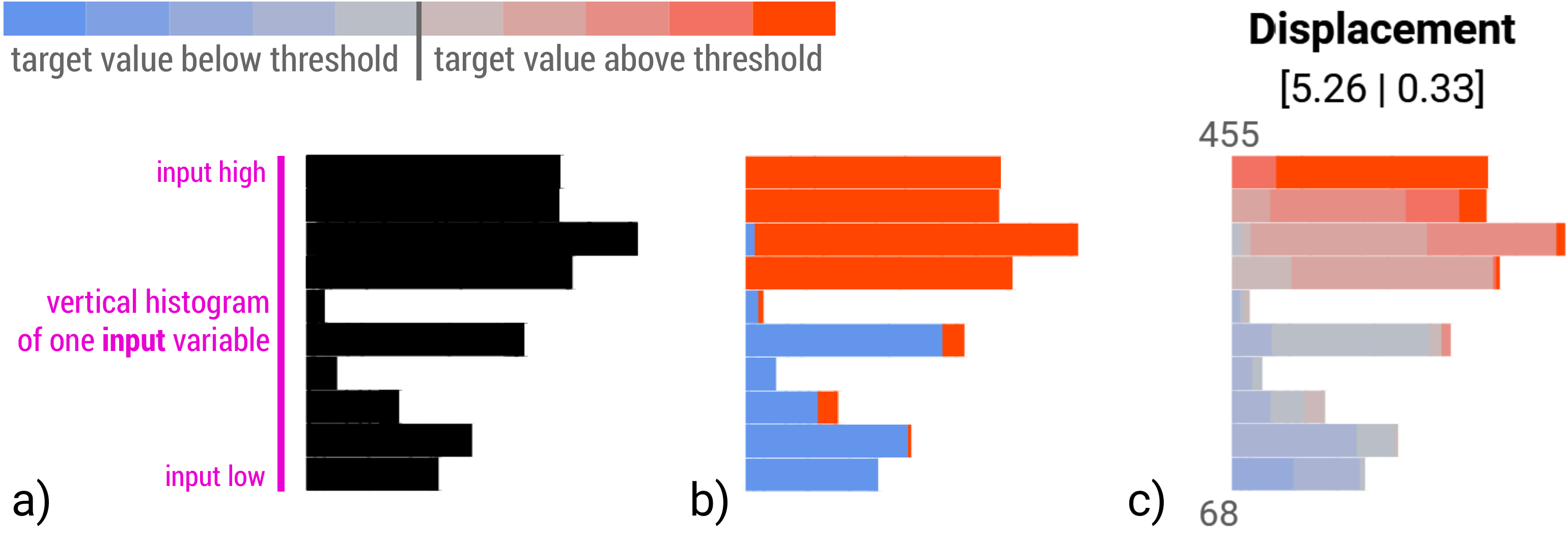}
  \caption{ Histogram of one particular input variable of a node: a) plain, b) two stacked bars per input bin for high- (red) and low-value (blue) cases, c) ten stacked bars per bin to visualize distribution of the target variable based on the input variable.}
  \label{fig:stacked-histogram}
\end{figure}

We want to visualize for which kind of input ranges the particular hidden node is mostly contributing.
To achieve this, we generate node-specific histograms of the input variables.
Given an input variable, we divide its range into equally distributed bins and count how many items in the data set fall within each bin, \emph{weighted} by the contribution of the node.
This results in a histogram of the variable that only incorporates data items where the node is contributing.
Figure~\ref{fig:stacked-histogram} a) shows how the resulting vertical histogram could look like.

In addition, we relate the input values to the value of the target variable by showing in how many cases the target value was above or below the threshold.
Hence, we count high- and low-value data items separately for each bin and display the resulting histogram as a stacked bar chart.
In other words, we build two separate histograms, one for those data items in the cluster with a high target value (red bars), and one for those with a low target value (blue bars).
Figure~\ref{fig:stacked-histogram} b) shows that for every data item where the particular node is contributing \emph{and} where the Displacement value falls into the top bin, the target value is always above the threshold (the bar is completely red).
Conversely, low Displacement values are associated with target values below the threshold, because the lower bars are nearly completely blue.

Instead of just distinguishing two cases (above vs. below the threshold), we enable users to increase the granularity to multiple bins.
This leads to histograms in a histogram.
For each bin of the input variable (rows), we visualize the distribution of the target variable (stacked bars within the row).
Analysts can choose the granularity of the histograms with two sliders at the bottom of the window (Figure~\ref{fig:cars-nodes}).
Figure~\ref{fig:stacked-histogram} c) shows the resulting visualization with ten input bins and ten target bins using colors from blue to red.
Within the respective cluster, \textit{Displacement} has a near-linear, positive correlation with the target value (horsepower), because, starting from the bottom, the bars change from blue-ish over gray to red-ish.

\subsubsection{Parallel Coordinate Plot}

\begin{figure}
  \centering
  \includegraphics[width=\linewidth]{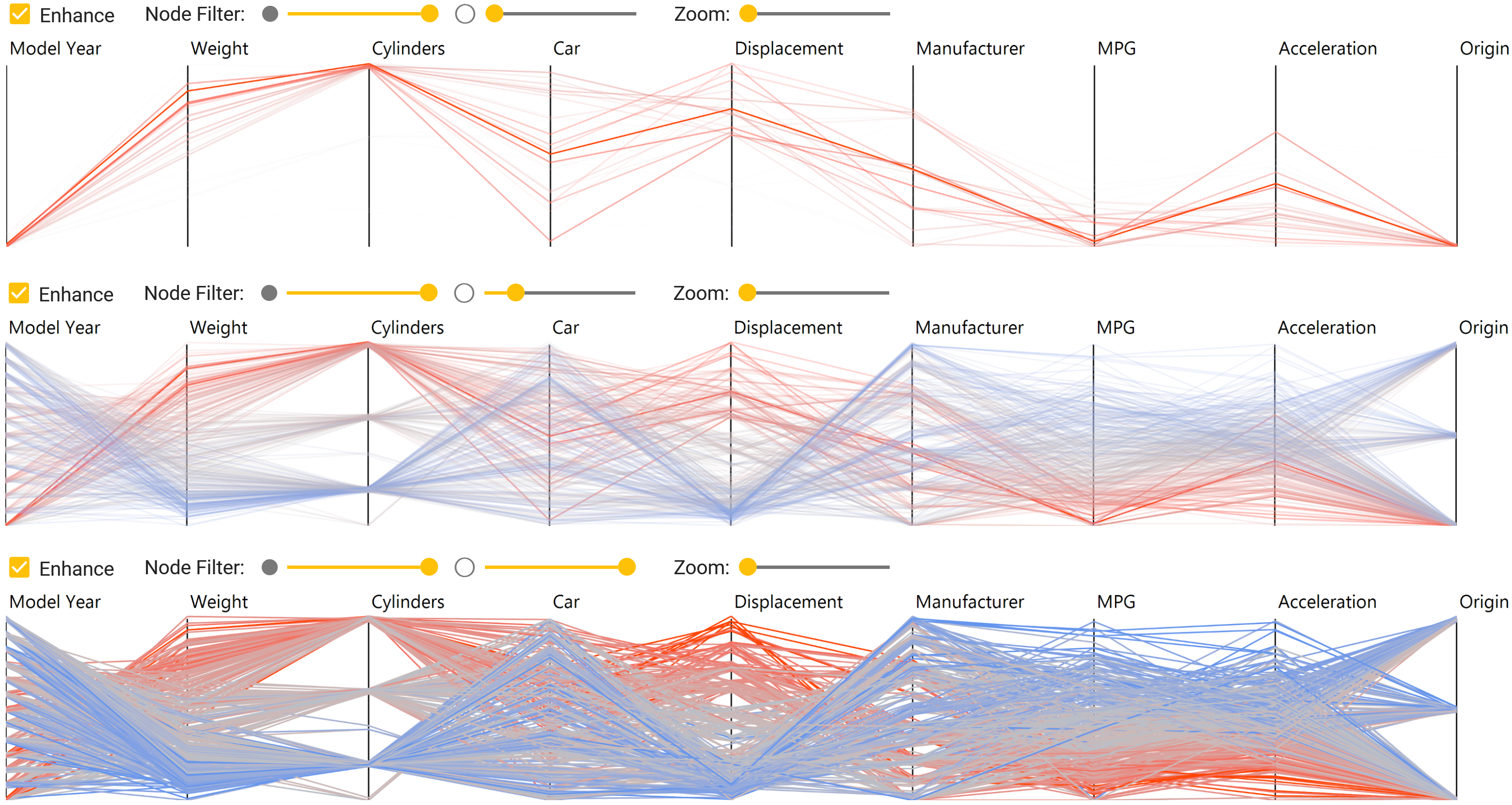}
  \caption{Node-specific PCPs with different filter settings. The columns are ordered by the importance of the inputs for the node. Top: Plot using only data items where this node is contributing. Middle: Plot includes remaining data items, but diminished. Bottom: Plot of all data items.}
  \label{fig:cars-pcp}
\end{figure}

We employ node-specific PCPs to let analysts explore \emph{parts} of the data set at a time and investigate relationships between the variables.
While the stacked histograms help to summarize which variables and ranges correlate with the target, we add the PCPs to allow for a detailed analysis how \emph{exactly} the input variables relate to each other for the given case represented by the node.
For instance, Node 1 in Figure~\ref{fig:cars-nodes} indicates that cars in our data set with high displacement and fast acceleration time have high horsepower.
The PCP shows that high displacement generally relates to low acceleration times, but in addition to the stacked histograms, it reveals that some cars with very high displacement only have average acceleration times.

One shortcoming of such plots is that it is often not obvious how the axes should be arranged, even though the order of the columns has a major influence on which patterns and correlations analysts can detect.
To tackle this challenge, we use the same order for the axes as for the stacked histograms, i.e., we arrange the columns according to their importance for the respective case represented by the the node.

Initially, we only draw data items where the node is contributing (filtered data), but this can be changed with two sliders above the plot, which we call \textit{node filters}.
The filtering mitigates occlusion issues that make it difficult to trace lines and recognize relationships for larger data sets.
The first slider (filled circle) determines the general opacity of the \emph{filtered data}, and the second (outline of a circle) the opacity of the \emph{remaining data}.
If both sliders are at zero, everything is transparent and nothing is shown.
If both are at maximum, the complete data set is displayed (but with the node-specific axis order).
Figure~\ref{fig:cars-pcp} shows three different filter settings of one node.
In all three cases, the first slider is at maximum, but the second slider changes from zero to intermediate to full.
This means that we slowly fade in the remaining data, which helps to relate the specific conditions to the entire data set.
In this example, the node focuses on heavy American (lowest position of \textit{Origin}) cars in the early 1970s (lowest position of \textit{Model Year}) that exhibit high horsepower (all lines are red).
However, if we only looked at the plot of the entire data set this pattern would be hard to detect visually.

%

\subsubsection{Range Filter}
\label{sec:rangeFilter}

\begin{figure}
  \centering
  \includegraphics[width=\linewidth]{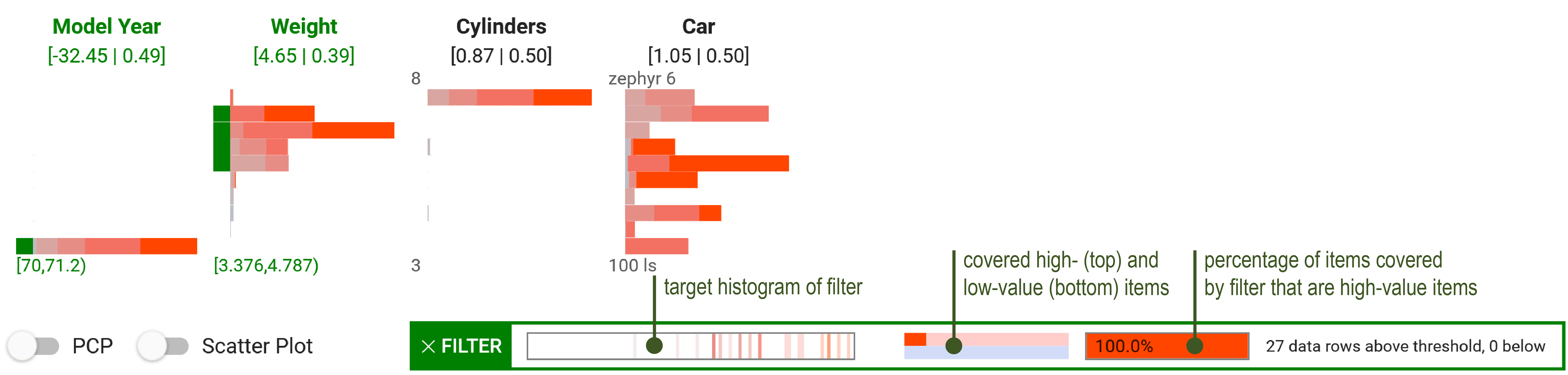}
  \caption{The analyst has selected several bars (green mark on the left) to build a simple range filter. The resulting histogram and statistics are shown below, which in this case indicate that heavier cars in the set built around the 1970s also exhibit above-average horsepower.}
  \label{fig:cars-rangefilter}
\end{figure}


Using the stacked histograms, analysts can build range filters to test hypotheses on interesting subsets of data items.
These filters reduce the set of data items to the ones having values in the specified ranges.
The stacked histograms in Figure~\ref{fig:cars-rangefilter} indicate that the respective node specializes in ancient, heavy, fast, American cars with many cylinders and high horsepower.
To validate whether most of the heavy cars around 1970 in the set have high horsepower, analysts can select the bottom bar of the first variable and the first few bars from the top of the \textit{Weight} variable by clicking on them.
Little green rectangles appear to the left of each selected bar, the name of the variable is marked in green, and a summary of the selected ranges (also in green) appears at the bottom of the histogram.
The resulting histogram and statistics of the filter appear below in the box with the green border.
Here, our range filter selects 27 cars and all exhibit horsepower above our threshold.
The small red bar in the middle of the green box (above the blue bar) shows that the filter covers roughly 10\% of all high-value cases.
Hence, cars in the data set from around 1970 which are heavier than average also have above-average horsepower.
Range filters always use the complete data set and do not depend on any node activation.
The node-specific visualizations, however, guide analysts to define such range filters.

\section{Evaluation}

In this section, we first show that our approach can detect and visualize correlations that involve more than two variables. Afterward, we will apply our method to real-world data sets to show its utility. Finally, we report on qualitative feedback on the suitability of the approach that we received from our domain expert.

\subsection{Identification of High-Value Cases}

\begin{figure}
  \centering
  \includegraphics[width=\linewidth]{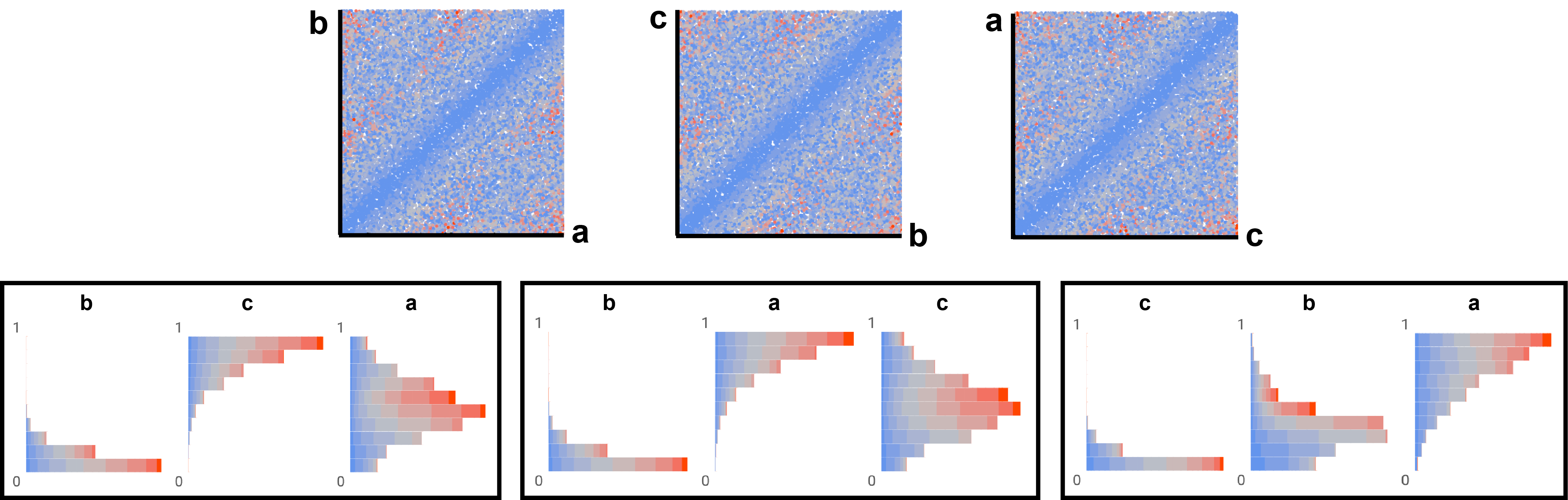}
  \caption{Comparison between scatter plots (top) and our approach (bottom) to visualize a four-dimensional artificial data set containing three different high-value patterns (different variable order due to importance ranking). The target variable is mapped to a color scale from blue to red.}
  \label{fig:comparison-3ind}
\end{figure}

We apply our approach to a synthetic data set to test its capability of identifying correlations between more than two variables.
The data set contains three input variables $a,b,c$ and 27,000 data rows that correlate with high values of a target variable under the following circumstances:

\begin{equation}
y(a,b,c) = \min (|a-b|,|b-c|,|c-a|)
\end{equation}

The target value reaches its maximum of $0.5$ only if one variable is $0$, one is $0.5$, and the third is $1$.
We uniformly sampled the input space $a,b,c \in [0,1]$ and generated $27,000$ rows.
We designed the target function $y(a,b,c)$ such that the resulting data set poses several challenges.
First, only about $12$\% of the items have target values above our threshold of $0.25$ and only 0.8\% above $0.4$, i.e., our machine learning model has few high-value samples to learn from.
Second, our target highly depends on the combination of all three variables, i.e., if we plot individual variables or pair of variables against our target value, we can hardly recognize any correlation.
Third, there are three different patterns that exhibit high target values, which would be occluded in a parallel coordinates plot.
We consider this data set to be a benchmark for testing our method under challenging conditions.

Figure~\ref{fig:comparison-3ind} shows the resulting stacked histograms of three hidden nodes of our trained model, and scatter plots of every combination of $a,b,c$ for comparison.
The target variable is mapped to a color scale from blue over gray to red in all plots.
From the scatter plots at the top, the analyst can conclude that the target is zero if any two variables have equal values, but it is not possible to recognize the three clusters of our ground-truth.
Conversely, the stacked histograms at the bottom reveal the three different patterns that lead to high values of $y$, namely the three permutations of $a,b,c$ where one variable is near $0$, one is near $1$, and the third is around the middle.
For instance, the first plot on the bottom-left indicates high target values (gray- and red-like bars) if $b$ is low, $c$ high, and $a$ around $0.5$.

\subsection{Use Cases on Real-World Data Sets}

VND supports the visual analysis of high-value cases in large multivariate data sets across a wide spectrum of use cases where the number of variables can be high.
Subsequently, we present two real-world use cases from different domains: IC chip testing and population surveys.

\subsubsection{Chip Testing Measurements}
\label{sec:chip-testing-results}

\begin{figure}
  \centering
  \includegraphics[width=\linewidth]{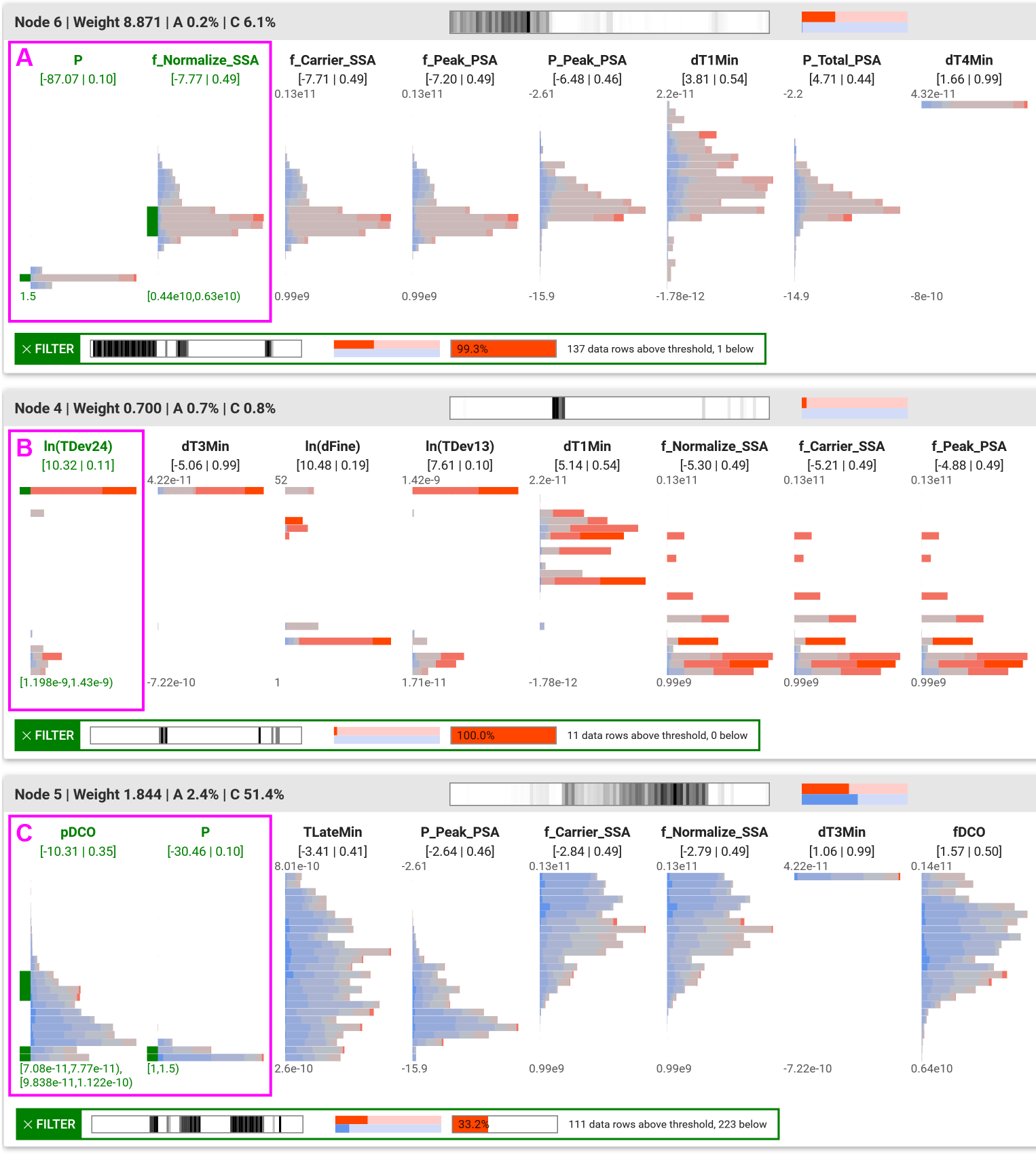}
  \caption{Decomposition of a chip testing data set. Higher target values (gray to red) correspond to faulty behavior of the clock device (jitter).
  }
  \label{fig:mb4vis-top3}
\end{figure}

Detecting and understanding complex input patterns that lead to errors in chip testing is a difficult endeavor.
We received a data set with measurements of a clock chip from our partner, a vendor of automatic test equipment.
The goal is to explore which conditions lead to high jitter as indicated by the target variable \textit{JTotal}.
Hence, our target variable is continuous, and the higher the value, the worse.
The set consists of 2049 data items and 43 variables including the target variable, and about 18\% of the items exhibit jitter above the desired threshold.

Figure~\ref{fig:mb4vis-top3} presents the resulting top three nodes after training.
These nodes mostly cover different parts of the data set according to the node coverage histograms (in black).
The output histograms (Figure~\ref{fig:teaser} shows the compact version on the very left) reveal that Node 6 and 5 mostly contribute to lower output values (jitter) around the threshold, whereas Node 4 specializes on few very high-value cases.
The Node 6-specific parallel coordinate plot is visible in Figure~\ref{fig:teaser} and depicts all data items where Node 6 is contributing.
This plot and the applied range filter in Figure~\ref{fig:mb4vis-top3} (A) show that a specific combination of the two most important variables for that first node almost always leads to target values above the threshold and can explain about half of all high-value cases.
For Node 4, the first variable is already enough to define 11 items with high jitter (B).
However, Node 5 shows a less conclusive picture (C).
The applied range filter and the red bar in the header indicate that the pattern represented by the node exhibits above-average jitter, but the node also contributes to several low-value cases.
This could mean that these conditions lead to unstable behavior which is sometimes in- and sometimes out-of-spec.

\subsubsection{AP VoteCast 2018}

The independent social research organization NORC at the University of Chicago conducted a survey of 138,929 registered US voters in 2018~\cite{Tompson2019}.  
To learn more about the voting behavior in the US presidential election 2016, we used the subset marked as \textit{nationally representative} with 4,913 registered voters, in which the participants were asked who they voted for in 2016.
1625 respondents (33.1\%) said they voted for Donald Trump, and 2129 (43.3\%) for Hillary Clinton.
We disabled variables associated with certain parties or politicians (e.g., whether the respondent likes politician \textit{X}) to gain insights into voting behaviors based on demographics and policy attitudes.
This reduced the number of variables to 67.

\begin{figure}
  \centering
  \includegraphics[width=\linewidth]{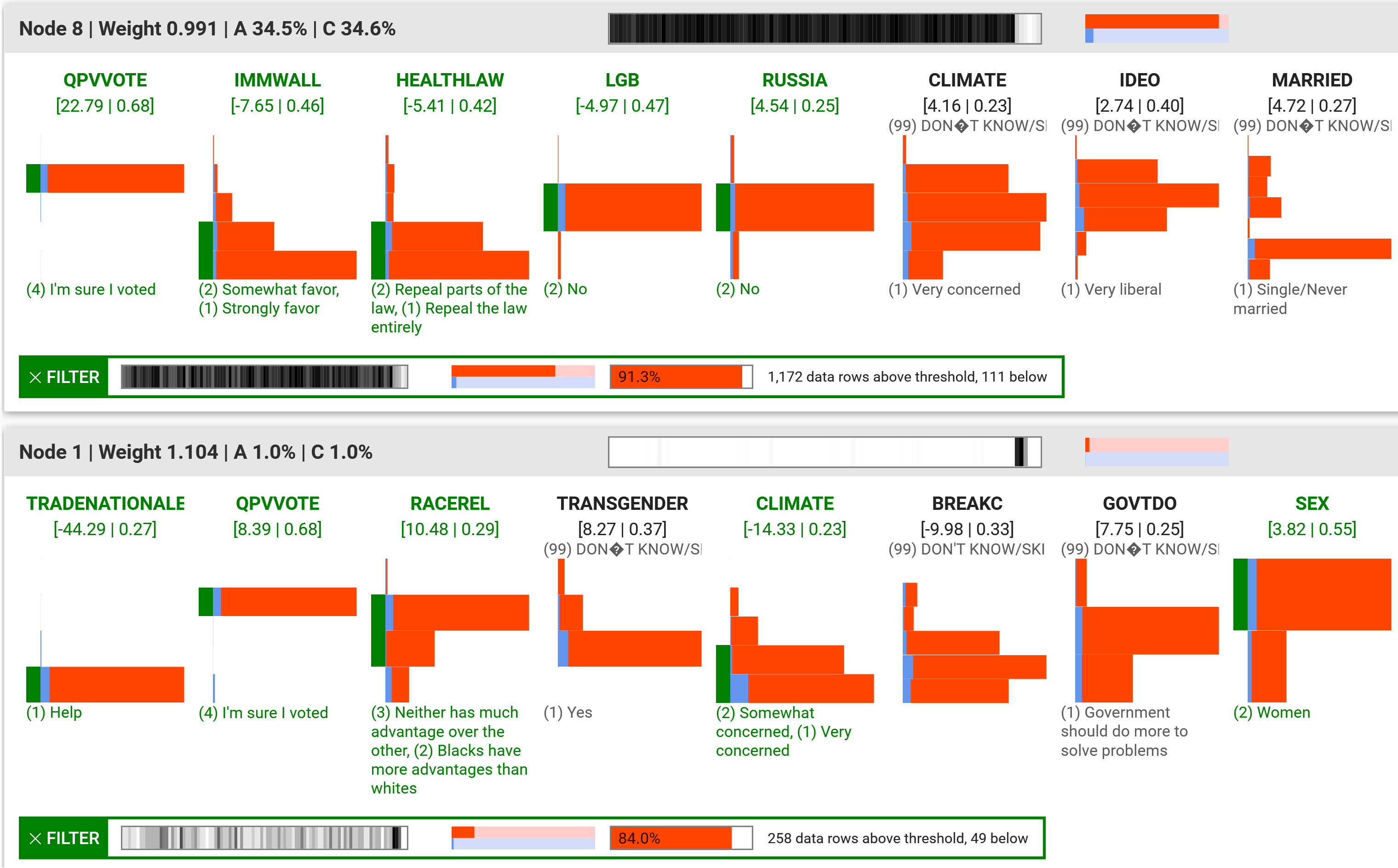}
  \caption{Decomposition of the AP VoteCast 2018 representative national survey data set comprising 4,913 respondents with 67 variables. Target variable is whether participants said they voted for Donald Trump in 2016.}
  \label{fig:trump-result-small}
\end{figure}

Figure~\ref{fig:trump-result-small} presents the result of our approach where we picked as target variable whether the participant voted for Donald Trump or not.
The top node shows that attitudes towards the border wall project and the Affordable Care Act (Obamacare) are strong predictors for the voting behavior of the participants.
Among those that are sure they voted in the presidential election (QPVVOTE), favor the border wall (IMMWALL), think that the Affordable Care Act should be at least partially repealed (HEALTHLAW), are not gay or bisexual (LGB), and do not think that the Trump campaign coordinated with Russia during the election (RUSSIA), 91.3\% said they voted for Trump, which represents 72.1\% of all Trump voters in the survey. 
Conversely, participants that oppose the border wall, think that the Care Act should be at least kept as it is, and are sure they voted in the election, 86\% voted for Clinton (see Appendix B for figures).

Node 1 in Figure~\ref{fig:trump-result-small} is interesting because it represents a cluster that only partially overlaps with Node 8.
The range filter reveals that 84\% of the female participants who are at least concerned about climate change (CLIMATE), but think that the trade policies of Trump's administration would help the national economy (TRADENATIONALECON), that Blacks have more or equally many advantages compared to Whites (RACEREL), and that they are sure they voted in the election, voted for Trump.
The percentage is significantly higher than the 29.7\% of Trump voters in the rest of the data set (Fisher's exact test, $p \ll 0.001$).
This is an interesting finding, because if we look at some of the variables individually, we observe different trends: 29.8\% of the female respondents, and 20.4\% of those at least somewhat concerned regarding climate change said they voted for Trump, which is in both cases less than the data set-wide average.

\begin{figure}
  \centering
  \includegraphics[width=\linewidth]{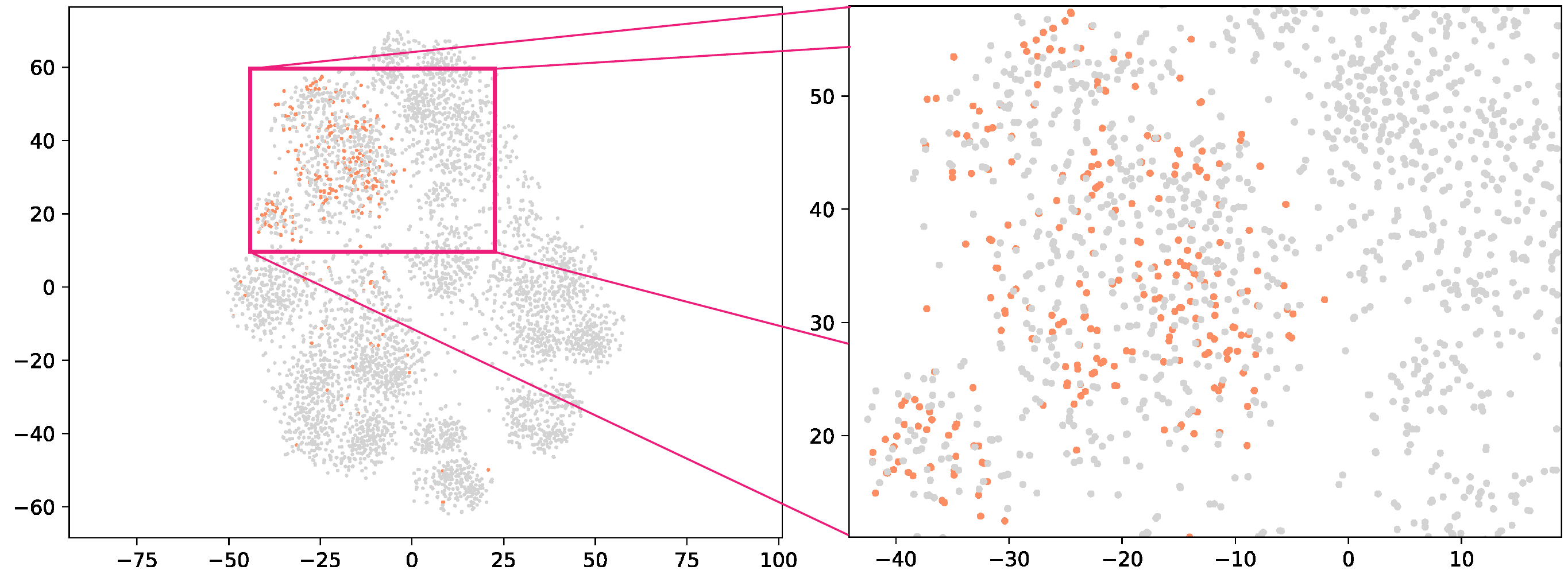}
  \caption{t-SNE projection of the VoteCast data. Items matching our range filter of Node 1 are colored orange.}
  \label{fig:pcaTsne}
\end{figure}

To investigate whether analysts would have found this insight using related methods, we applied PCA~\cite{Abdi2010}, t-SNE~\cite{VanDerMaaten2008} and UMAP~\cite{McInnes2018} to the exact same data set.
It turns out that our cluster of interest does not form a visual cluster in any of the projections (Figure~\ref{fig:pcaTsne} depicts the result for t-SNE).
Hence, with the projections it would be difficult for analysts to detect and explore a cluster comparable to the one we found with our approach (see Appendix B for the detailed analysis).
This use case also shows that a particular combination of several variables can exhibit much stronger (and sometimes opposite) correlations with the target variable than individual variables or pairs, while at the same time still covering a large proportion of interesting high-value cases.

\subsection{Qualitative Feedback}

We presented a domain expert and research fellow of the world's largest provider of automatic test equipment an earlier version of our approach and collected preliminary feedback.
We introduced the system and showed the results of the data set the company provided (Section~\ref{sec:chip-testing-results}).
The expert was impressed that our approach was capable to detect high-value conditions and that it supports the exploration of highly interesting combinations of variables.
He was confident that this would greatly support the engineers in finding problematic conditions.
However, he noted that there is a learning curve as it takes some time to understand the stacked histogram visualizations.

The expert made several suggestions on how to improve the utility of our approach.
It should be possible to filter by value range of certain variables (not just by nodes), the target threshold should be customizable, and it should be possible to remove or ignore data items based on clear correlations to focus on the remaining, less-understood parts of the data set.
Based on the feedback, we implemented the range filter and made the target threshold adjustable.
In addition, we developed the node coverage histograms.

\section{Discussion}

Visual Neural Decomposition (VND) can extract high-value conditions in multivariate data sets that contain many variables and it guides analysts to explore non-linear multi-variable relationships.
Our use cases show that the approach can also reveal significant relationships that are difficult to find with related methods, e.g., approaches applying dimensionality reduction techniques or linear models.
With our novel approach, analysts do not need to iteratively select one or two variables at a time to start the process.
Instead, we separately visualize groups of conditions based on the original dimensions that often have a semantic meaning in multivariate data sets.
Additionally, the automatic analysis does not only cluster the results, but it also provides a ranking of the most important variables of each cluster.
This helps users to grasp the meaning of each cluster.
The provided interaction possibilities such as the range filter or node-specific parallel coordinate plots enable analysts to validate their hypotheses and build trust in what they see.

Our proposed technique has certain limitations.
While it scales to hundreds of variables, it was not designed for the analysis of very high-dimensional data sets, for instance, images with millions of pixels.
However, in this case, it is often not helpful to explain certain conditions based on single pixel values.
A shortcoming is that analysts have to define hyper-parameters, e.g., the number of nodes, and each run can lead to a slightly different output.
Nevertheless, our analysis (Section~\ref{sec:regularization}) shows that the method is not very sensitive to the specified parameters.
Generally, more nodes and longer training are better, but also computationally more expensive.
In our use cases, the training only took seconds, though.
The regularization constrains the solution space, but there can still be several equally valid definitions of clusters, particularly if the variables are not independent of each other.
The goal of VND is to visually explain high-value cases, hence, if one input is already enough to perfectly predict the target then VND probably will not extract additional (somewhat redundant) cases.
Finally, it is important to realize that our approach visualizes interesting conditions that appear \emph{in the data}.
If analysts want to generalize such findings, e.g., whether an observed voting behavior applies to the general population, they still need to perform statistical tests for significance.

\section{Conclusion and Future Work}

We proposed a novel method to visually explain multivariate data sets using neural networks.
The idea is to decompose the data set into components as represented by the hidden nodes and provide node-specific views of the data set.
For this, we developed an integrated approach to configure the data, train the model, visualize the results, and offer interaction mechanisms to let analysts validate hypotheses.

In the future, we aim to further improve the analysis workflow.
We would like to support negative filters, i.e., analysts can select items based on nodes or value ranges that should be ignored in a subsequent run to encourage explanations for the remaining, less understood parts of the data set.
Letting users define unions or intersections of nodes and range filters could help to quantify how many high-value cases are covered in total.
Finally, providing an additional mode to align variables with those of a reference node would enable a detailed comparison of interesting subsets.

\acknowledgments{
This research was partially supported by Advantest as part of the Graduate School ‘Intelligent Methods for Test and Reliability’ (GS-IMTR) at the University of Stuttgart, and by the German Science Foundation (DFG) as part of the project ‘VAOST’ (project number 392087235). In particular, we would like to thank Jochen Rivoir from Advantest for giving us access to the chip test data set and for his valuable feedback.
}

\bibliographystyle{abbrv-doi}

\bibliography{vnca-paper}
\end{document}